\documentclass{article}
\PassOptionsToPackage{numbers,square}{natbib}

\usepackage[main,final]{neurips_2026}

\usepackage[utf8]{inputenc}
\usepackage[T1]{fontenc}
\usepackage{hyperref}
\usepackage{url}
\usepackage{booktabs}
\usepackage{amsfonts}
\usepackage{amsmath}
\usepackage{amssymb}
\usepackage{mathtools}
\usepackage{amsthm}
\usepackage{nicefrac}
\usepackage{microtype}
\usepackage{xcolor}
\usepackage{graphicx}
\usepackage{subcaption}
\usepackage{multirow}
\usepackage{pifont}     
\usepackage{colortbl}   
\usepackage{algorithm}
\usepackage{algorithmic}
\usepackage{bm}
\usepackage[capitalize,noabbrev]{cleveref}

\newcommand{\cmark}{\ding{51}}
\newcommand{\xmark}{\ding{55}}

\theoremstyle{plain}

\theoremstyle{definition}

\theoremstyle{remark}

\newcommand{\divmoe}{\textsc{DivMoE}}
\newcommand*{\divmoecodeurl}{}
\newcommand{\divmoecodelink}{%
  \ifx\divmoecodeurl\empty
    \texttt{[GitHub link pending]}%
  \else
    \href{\divmoecodeurl}{GitHub repository}%
  \fi
}

\title{DivMoE: Fine-Grained MoE Upcycling via\\Cross-Domain Expert Composition}

\author{%
  Yuxuan Lou\textsuperscript{1}\thanks{Equal contribution.}\quad
  Kai Yang\textsuperscript{2}\footnotemark[1]\quad
  Geng Zhang\textsuperscript{1}\quad
  Yong Liu\textsuperscript{1}\quad
  Yang You\textsuperscript{1}\thanks{Corresponding author: \href{mailto:youy@comp.nus.edu.sg}{\texttt{youy@comp.nus.edu.sg}}.}\\
  \textsuperscript{1}School of Computing, National University of Singapore, Singapore\\
  \textsuperscript{2}Department of Computer Science and Engineering,\\
  Shanghai Jiao Tong University, Shanghai, China
}
\hypersetup{
  hidelinks,
  pdfauthor={Yuxuan Lou, Kai Yang, Geng Zhang, Yong Liu, Yang You},
  pdftitle={DivMoE: Fine-Grained MoE Upcycling via Cross-Domain Expert Composition}
}
\PackageWarningNoLine{divmoe-final}{Two author affiliations, funding/disclosures, and GitHub URL are pending. Do not submit this draft}

\begin{document}

\begingroup
\makeatletter
\renewcommand{\@fnsymbol}[1]{%
  \ifcase#1\or\ensuremath{\dagger}\or\mbox{\ding{41}}\else\@ctrerr\fi
}
\makeatother
\maketitle
\endgroup
\setcounter{footnote}{0}

\begin{abstract}
Mixture-of-Experts (MoE) architectures have become essential for scaling large language models, with recent work demonstrating the benefits of \emph{fine-grained} expert designs. Training such models from scratch is expensive, and \emph{sparse upcycling} from pre-trained dense models is an attractive alternative. However, we identify a structural pathology of fine-grained upcycling: when fine-grained experts are derived from a single source model, naive routing collapses and downstream accuracy drops to near-random (e.g., on Qwen3-1.7B, Drop-Upcycling-fine-grained reaches only 23.2\% average accuracy across 15 benchmarks, essentially matching from-scratch training at 22.2\%, while the same method's coarse-grained variant reaches 50.2\%). We propose \divmoe{}, the first framework achieving fine-grained MoE Upcycling with structurally-balanced routing. \divmoe{} introduces \emph{domain-specialized fine-grained expert initialization}, deriving experts from dense models that have undergone domain-adaptive continual pre-training, and \emph{diversity-constrained routing}, a hard structural constraint guaranteeing that each token activates experts from distinct domain groups. Across two base models and 15 benchmarks, \divmoe{} consistently outperforms six upcycling baselines (55.6\% vs.\ 51.6\% for the strongest baseline on Qwen3-1.7B) and \emph{strictly improves over the dense base model on every benchmark} after Stage~2 continual pre-training---closing the regression gap that has plagued prior fine-grained upcycling. After supervised fine-tuning on a public reasoning mixture, our 12B-parameter \divmoe{} model matches Moonlight-MoE (16B) at 64.5\% average accuracy while outperforming a controlled NVIDIA-Upcycling baseline by 6.1 percentage points.
\end{abstract}

\section{Introduction}
\label{sec:introduction}
Mixture-of-Experts (MoE) architectures have emerged as a cornerstone for building state-of-the-art large language models, enabling substantial capacity expansion while maintaining computational efficiency through sparse activation~\citep{shazeer2017outrageously,fedus2022switch}. Recent advances have demonstrated the effectiveness of \emph{fine-grained} MoE designs, where each feed-forward network (FFN) layer is partitioned into many small experts rather than a few large ones~\citep{dai2024deepseekmoe,liu2024deepseekv2} by enabling more flexible routing and finer specialization, which contributes to the strong performance of models such as DeepSeek-V2 and Mixtral~\citep{jiang2024mixtral}.

Training such models from scratch demands substantial compute, motivating \emph{sparse upcycling}~\citep{komatsuzaki2022sparse}, which converts pre-trained dense checkpoints into MoE architectures while inheriting their learned representations. However, we observe that fine-grained upcycling is fundamentally fragile: when fine-grained experts are derived from a single dense source, downstream accuracy collapses to near-random. On Qwen3-1.7B-Base evaluated across 15 benchmarks, Drop-Upcycling-fine-grained~\citep{chen2024dropupcycling} reaches only 23.2\% average accuracy (comparable to from-scratch training, 22.2\%), versus 50.2\% for the same method's coarse-grained variant. NVIDIA Upcycling~\citep{he2024upcycling}, which preserves dense-equivalence by copying a single source, recovers the gap (51.1\%) but yields homogeneous expert initial scores that suppress routing diversity. The two existing remedies are unsatisfactory: identical weight copying limits routing diversity by construction, while random re-initialization sacrifices pre-trained knowledge.

Our central observation is that no prior method simultaneously occupies three design axes that the experiments below show are \emph{jointly necessary}: (a)~domain-specialized expert initialization, (b)~fine-grained partitioning, and (c)~a structural mechanism that prevents same-source co-activation. Branch-Train-Mix (BTX)~\citep{sukhbaatar2024branch} addresses (a) but uses coarse experts and standard top-$k$ routing; NVIDIA Upcycling addresses (b); Drop-Upcycling targets diversity through randomness, sacrificing (a) in a different sense. The concurrent FlexOLMo~\citep{shi2025flexolmo} also constructs experts from domain specialists, but for the orthogonal goal of flexible data governance: it uses coarse experts, no joint training, and no routing constraint.

We propose \divmoe{}, the first framework that achieves fine-grained MoE Upcycling, which includes two components:
(1)~\emph{Domain-specialized fine-grained expert initialization.} We derive experts from $n$ dense models that have undergone \emph{domain-adaptive continual pre-training} on public corpora (mathematics, code, science, commonsense), then vertically slice each specialist's FFN into $m$ shards, yielding $N=n\times m$ fine-grained experts organized into $n$ domain groups. This preserves all pre-trained knowledge while injecting structured diversity into the routing scores at initialization.
(2)~\emph{Diversity-constrained routing.} We enforce a hard structural constraint that each token selects at most one expert per domain group. Unlike soft auxiliary losses~\citep{fedus2022switch,zoph2022stmoe} that trade routing balance against task accuracy, the diversity constraint operates outside the loss landscape and provides a guarantee regardless of gradient magnitudes.

Empirically, \divmoe{} consistently outperforms six upcycling baselines across 15 benchmarks on two base models (Qwen3-1.7B-Base, Llama3.2-1B). With Stage~2 continual pre-training on the public  Dolmino-Mix-1124 corpus~\citep{olmo2025dolmino}, \divmoe{} \emph{strictly improves over the dense base model on the majority of benchmarks}, eliminating the post-CPT regression that prior upcycling pipelines exhibit. Crucially, the structural constraint improves performance on every domain-specific benchmark (HumanEval +1.8\%, GSM8K +2.1\%, MATH +1.4\%, MBPP +2.3\%), refuting the intuition that ``coding tokens should only use coding experts.'' After Stage~3 supervised fine-tuning on a public reasoning mixture, our 12B-parameter \divmoe{} model matches Moonlight-MoE (16B) at 64.5\% and outperforms a controlled NVIDIA-Upcycling baseline by 6.1 percentage points.

The contributions of our work are summarized below:
\begin{itemize}
  \item \textbf{Observation on a fine-grained upcycling collapse pathology.} We show that combining fine-grained expert partitioning with single-source initialization leads to routing collapse and severe performance degradation.
  \item \textbf{The first fine-grained MoE Upcycling framework.} We propose \divmoe{}, which combines domain-specialized fine-grained expert initialization with a hard diversity-constrained routing mechanism.
  \item \textbf{Comprehensive experiments with strong downstream performance.} 
  Across two base models, three model scales (1B, 1.7B, 4B base $\to$ 3.2B, 5.4B, 12B MoE), fifteen benchmarks, and six upcycling baselines, all using public corpora, \divmoe{} consistently outperforms prior upcycling methods and closes the regression gap of previous fine-grained upcycling approaches.
\end{itemize}

\section{Related Work}
\label{sec:related_work}

\paragraph{Mixture-of-Experts in LLMs.}
MoE architectures scale model capacity efficiently through conditional computation~\citep{jacobs1991adaptive,jordan1994hierarchical,shazeer2017outrageously}. GShard~\citep{lepikhin2021gshard} and Switch Transformer~\citep{fedus2022switch} scaled this paradigm to trillion-parameter models with top-$k$ routing and load-balancing losses. Mixtral~\citep{jiang2024mixtral} matches dense quality at reduced cost, while DeepSeek-MoE~\citep{dai2024deepseekmoe} and DeepSeek-V2~\citep{liu2024deepseekv2} introduced \emph{fine-grained expert segmentation} with shared experts. OLMoE~\citep{muennighoff2024olmoe} provides open MoE models and data. ST-MoE~\citep{zoph2022stmoe} introduced the router z-loss for stable routing. \divmoe{} extends this fine-grained trend by deriving fine-grained experts from \emph{multiple} domain-specialized dense models and enforcing routing diversity structurally.

\paragraph{Dense-to-MoE upcycling.}
Sparse Upcycling~\citep{komatsuzaki2022sparse} initializes all experts from copies of the same dense FFN. Branch-Train-Mix~\citep{sukhbaatar2024branch} (BTX, distinct from Branch-Train-Merge~\citep{li2022branchtrain}) trains separate dense ``branch'' models on different data slices, then assembles them into a coarse-grained MoE in which each expert is one full FFN. \citet{he2024upcycling} extend dense-equivalent initialization to fine-grained experts, but derive all experts from a single source model, providing no initial routing-score diversity. Drop-Upcycling~\citep{chen2024dropupcycling} introduces diversity via random re-initialization of expert subsets, at the cost of partial knowledge loss; we will show that its fine-grained variant collapses. The concurrent FlexOLMo~\citep{shi2025flexolmo} also constructs experts from domain specialists, but for the orthogonal goal of flexible data governance: experts train independently with no joint pre-training, the granularity is coarse, and no routing constraint is imposed.

\paragraph{Routing collapse and load balancing.}
Top-$k$ routing is prone to \emph{routing collapse}, in which a few experts dominate selection regardless of input~\citep{fedus2022switch}. Standard remedies are auxiliary load-balancing losses~\citep{shazeer2017outrageously,fedus2022switch} or expert-choice routing~\citep{zhou2022mixtureofexperts}. The router z-loss~\citep{zoph2022stmoe} penalizes large router logits to stabilize training. All of these are \emph{soft} mechanisms: they add gradients to the routing logits without changing what selections are reachable. \divmoe{}'s diversity-constrained routing is a complementary \emph{hard} mechanism whose guarantee is independent of gradient magnitudes; we retain a small auxiliary load-balancing loss (\Cref{sec:routing}) only to encourage utilization \emph{within} domain groups.

\paragraph{Continual pre-training.}
Multi-stage training pipelines that interleave continual pre-training with task-specific fine-tuning are now standard~\citep{ibrahim2024simple,gupta2023continual}. The choice of CPT data has a large effect on benchmark performance, with mismatches between the base model's original pre-training distribution and the CPT mixture causing transient regressions on specific tasks. We use Dolmino-Mix-1124~\citep{olmo2025dolmino}, a public mid-training corpus curated for the OLMo-2 family, which is high-enough quality that all upcycling baselines (including ours) post-CPT improve over their dense bases on the majority of benchmarks.

\paragraph{Positioning.}
\Cref{tab:design-space} summarizes the design space. \divmoe{} is the first method to achieve fine-grained MoE Upcycling, with the mechanism occupying all three columns simultaneously.

\begin{table}[t]
\caption{Design-space of dense-to-MoE upcycling. \cmark indicates the method occupies the design choice. Only \divmoe{} occupies all three columns; the empirical consequence is that fine-grained partitioning without domain diversity collapses (\Cref{sec:upcycling_comparison}).}
\label{tab:design-space}
\centering
\small
\begin{tabular}{lccc}
\toprule
\textbf{Method} & \textbf{Domain-specialized} & \textbf{Fine-grained} & \textbf{Structural routing} \\
 & \textbf{experts} & \textbf{partitioning} & \textbf{constraint} \\
\midrule
Sparse Upcycling~\citep{komatsuzaki2022sparse}      & \xmark & \xmark & \xmark \\
Branch-Train-Mix~\citep{sukhbaatar2024branch}        & \cmark & \xmark & \xmark \\
NVIDIA Upcycling~\citep{he2024upcycling}             & \xmark & \cmark & \xmark \\
Drop-Upcycling~\citep{chen2024dropupcycling}         & \xmark & \xmark & \xmark \\
Drop-Upcycling-fine-grained                          & \xmark & \cmark & \xmark \\
FlexOLMo~\citep{shi2025flexolmo} (concurrent)        & \cmark & \xmark & \xmark \\
\rowcolor{gray!15}
\textsc{DivMoE} (ours)                               & \cmark & \cmark & \cmark \\
\bottomrule
\end{tabular}
\end{table}

\section{DivMoE}
\label{sec:method}
\begin{figure}[t]
\centering
\includegraphics[width=\linewidth]{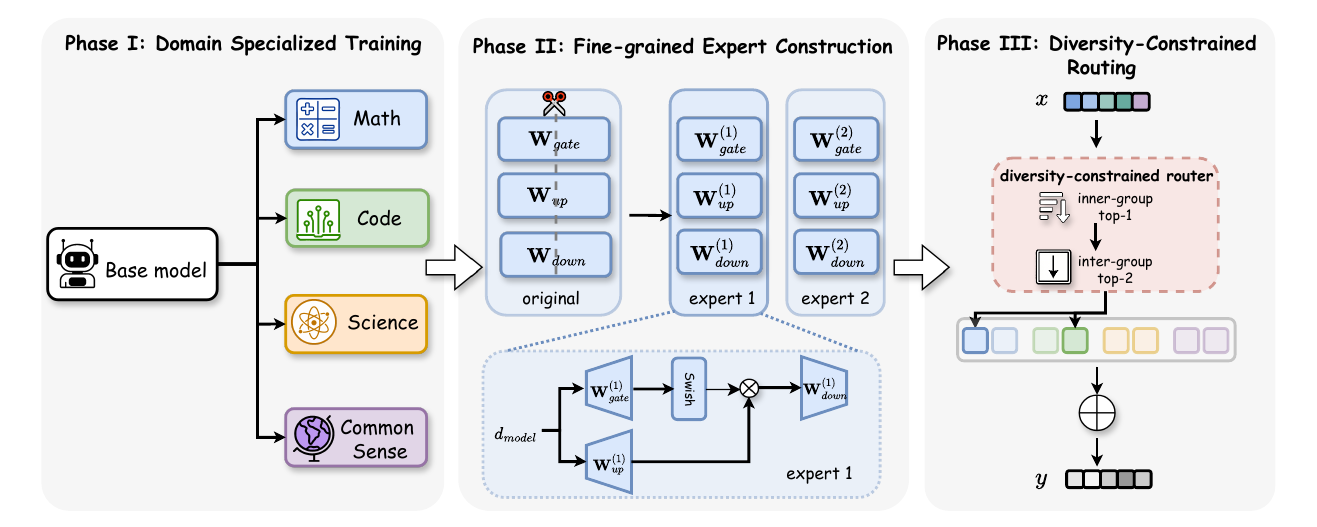}
\caption{\divmoe{} overview. \textbf{Left:} a single base model is domain-adaptively pre-trained into $n=4$ specialists. \textbf{Middle:} each specialist's FFN is vertically sliced into $m=2$ shards, yielding $N=8$ experts in $n=4$ domain groups. \textbf{Right:} the router selects the top-1 expert within each group, then top-$k$ groups, guaranteeing $k$ experts from $k$ distinct domains.}
\label{fig:architecture}
\end{figure}

\divmoe{} upcycles dense language models into fine-grained MoE architectures through two innovations: (1)~\emph{domain-specialized fine-grained expert initialization} (\Cref{sec:initialization}) that derives experts from $n$ dense models domain-adaptively pre-trained on distinct corpora, and (2)~\emph{diversity-constrained routing} (\Cref{sec:routing}) that structurally enforces cross-domain expert selection. \Cref{fig:architecture} provides an overview.

Given a dense transformer with SwiGLU FFN $\text{FFN}(\mathbf{x}) = \mathbf{W}_{\text{down}} \cdot \sigma(\mathbf{W}_{\text{gate}} \mathbf{x}) \odot (\mathbf{W}_{\text{up}} \mathbf{x})$, where $\mathbf{W}_{\text{gate}}, \mathbf{W}_{\text{up}} \in \mathbb{R}^{d_{\text{ff}} \times d}$ and $\mathbf{W}_{\text{down}} \in \mathbb{R}^{d \times d_{\text{ff}}}$, we convert each FFN into an MoE layer. Unlike prior work that derives experts from a single model~\citep{komatsuzaki2022sparse,he2024upcycling}, \divmoe{} first trains $n$ \emph{domain specialists} $\{\mathcal{M}_1, \ldots, \mathcal{M}_n\}$ from the base model, then slices each specialist's FFN into $m$ fine-grained experts. With $n=4$, $m=2$, and $k=2$, \divmoe{} doubles total capacity while activating half the dense FFN parameters.

\subsection{Domain-Specialized Fine-Grained Expert Initialization}
\label{sec:initialization}

\paragraph{Domain specialists.}
Starting from base model $\mathcal{M}_{\text{base}}$, we apply \emph{domain-adaptive continual pre-training} (next-token-prediction loss on full sequences, not instruction-style SFT) on public domain-specific subsets to obtain $n$ specialists $\{\mathcal{M}_{\text{math}}, \mathcal{M}_{\text{code}}, \mathcal{M}_{\text{science}}, \mathcal{M}_{\text{common}}\}$. Each specialist encodes distinct domain knowledge while retaining general capabilities, in contrast to identical copies~\citep{komatsuzaki2022sparse} or random perturbations~\citep{chen2024dropupcycling}.

\paragraph{Expert construction.}
Each specialist's FFN weights are vertically sliced into $m$ shards:
\begin{align}
\mathbf{W}_{\text{gate},i} &\rightarrow \{\mathbf{W}_{\text{gate},i}^{(1)}, \ldots, \mathbf{W}_{\text{gate},i}^{(m)}\}, \\
\mathbf{W}_{\text{up},i} &\rightarrow \{\mathbf{W}_{\text{up},i}^{(1)}, \ldots, \mathbf{W}_{\text{up},i}^{(m)}\}, \\
\mathbf{W}_{\text{down},i} &\rightarrow \{\mathbf{W}_{\text{down},i}^{(1)}, \ldots, \mathbf{W}_{\text{down},i}^{(m)}\},
\end{align}
where expert $(i,j)$ has dimensions $(d_{\text{ff}}/m) \times d$ for gate/up and $d \times (d_{\text{ff}}/m)$ for down projections. Indices $i\in\{1,\ldots,n\}$ and $j\in\{1,\ldots,m\}$ identify the specialist (domain group) and shard, respectively.

\paragraph{Dense-equivalent initialization.}
Following \citet{he2024upcycling}, we ensure stable initialization via \emph{virtual groups}: selecting one shard $j$ from any specialist reconstructs a complete FFN. Decomposing each specialist as $\mathbf{W}_i = \mathbf{W}_{\text{base}} + \boldsymbol{\Delta}_i$ where $\boldsymbol{\Delta}_i$ captures domain adaptations,
\begin{equation}
\sum_{j=1}^{m} \text{Expert}_{d_j}^{(j)}(\mathbf{x}) \approx \text{FFN}_{\text{base}}(\mathbf{x}),
\end{equation}
preserving the base model's output distribution at initialization while introducing structured diversity through the $\boldsymbol{\Delta}_i$.

\paragraph{Magnitude correction.}
We scale expert weights by $\gamma$ to compensate for softmax attenuation of router scores~\citep{he2024upcycling}:
\begin{equation}
\mathbf{W}_{\text{expert}} = \gamma \cdot \mathbf{W}_{\text{slice}}, \quad \gamma = \sqrt[3]{\frac{E \cdot G^2}{k}},
\end{equation}
where $E$ is the expansion factor, $G$ is granularity, and $k$ is the active-expert count.

\paragraph{Non-expert components.}
Embeddings, attention, and normalization layers are initialized as the mean of the base model and all specialists:
\begin{equation}
\mathbf{W}_{\text{shared}} = \frac{1}{n+1} \left( \mathbf{W}_{\text{base}} + \sum_{i=1}^{n} \mathbf{W}_i \right).
\label{eq:nonexpert}
\end{equation}
We ablate this choice in \Cref{app:attention_init}.

\subsection{Diversity-Constrained Routing}
\label{sec:routing}

\begin{figure}[t]
\centering
\includegraphics[width=1.0\linewidth]{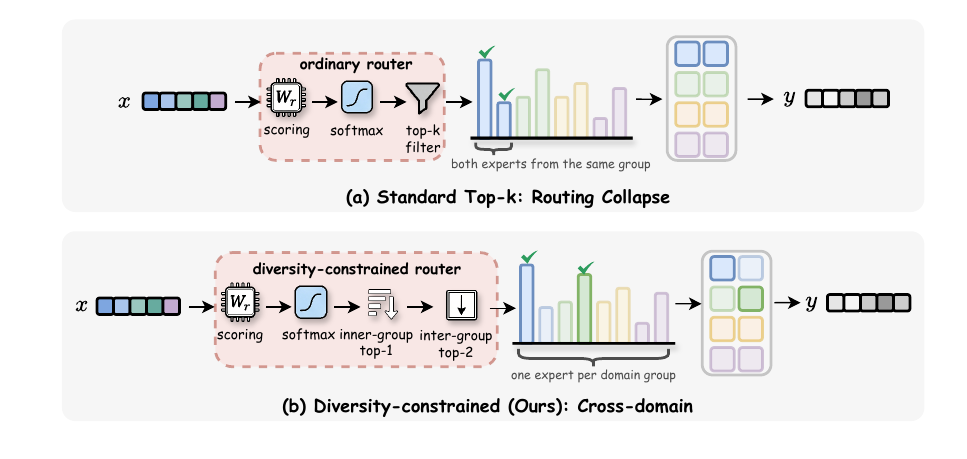}
\caption{Routing comparison. (a)~Standard top-$k$ may select both experts from one domain. (b)~Diversity-constrained routing guarantees cross-domain selection.}
\label{fig:routing}
\end{figure}

Standard MoE routing computes scores $\mathbf{s} = \mathbf{W}_r \mathbf{x} \in \mathbb{R}^N$ and selects $\mathcal{S} = \text{TopK}(\text{Softmax}(\mathbf{s}), k)$. When experts are partitioned into $n$ domain groups $\{\mathcal{D}_1,\ldots,\mathcal{D}_n\}$ as in \Cref{sec:initialization}, this top-$k$ rule admits within-group concentration: tokens may route to multiple experts in a single group, leaving other groups inactive (\emph{routing collapse}, \Cref{fig:routing}a). To prevent this, we impose a \emph{diversity constraint} that each token activates at most one expert per group:
\begin{equation}
\forall i \in \{1, \ldots, n\}: \quad |\mathcal{S} \cap \mathcal{D}_i| \leq 1,
\label{eq:diversity_constraint}
\end{equation}
combined with $k \leq n$, \Cref{eq:diversity_constraint} guarantees that the $k$ selected experts originate from $k$ distinct domain groups (\Cref{fig:routing}b).

The constraint is intentionally structural rather than gradient-based. Conventional remedies for routing imbalance---auxiliary load-balancing losses~\citep{fedus2022switch} and router z-losses~\citep{zoph2022stmoe}---are \emph{soft}: they shape routing through gradient pressure that competes with the language-modeling loss, and they offer no guarantee on what selections are reachable. In contrast, \Cref{eq:diversity_constraint} eliminates same-group co-activation by construction, independently of gradient magnitudes. The structural choice is also justified by the initialization: experts within a group are slices of the same domain specialist and encode similar transformations, $\text{Expert}_{i,j}(\mathbf{x}) \approx \text{Expert}_{i,j'}(\mathbf{x})$ for $j,j' \in \{1,\ldots,m\}$, so selecting more than one is largely redundant.

A potential concern is that obligatory cross-domain composition may waste capacity on ostensibly domain-specific tasks---e.g., forcing a coding token to also draw from math, science, or commonsense experts. We argue, and later confirm empirically (\Cref{sec:ablation_diversity}), that the constraint operates at a finer granularity than the prompt: individual tokens within a coding prompt serve heterogeneous functional roles (natural-language description, algorithmic reasoning, syntactic generation) and benefit from complementary expertise. Critically, the constraint does not equalize routing weights; the softmax scores still reflect input distribution, so domain-relevant experts continue to receive higher weights and domain emphasis is preserved (\Cref{fig:routing_patterns}).

\Cref{alg:routing} realizes the constraint in two deterministic stages: for each domain group $i$, select the highest-scoring expert $j_i^* = \arg\max_j s_{i,j}$; then select the top-$k$ groups by their representative scores $s_i^* = s_{i,j_i^*}$. This procedure adds no parameters or hyperparameters, integrates with standard MoE infrastructure, and---together with a small auxiliary load-balancing loss ($\alpha=0.01$) retained for \emph{within}-group balance---guarantees $k$ experts drawn from $k$ distinct domains.

\begin{algorithm}[t]
\caption{Diversity-Constrained Routing}
\label{alg:routing}
\begin{algorithmic}
\STATE {\bfseries Input:} hidden state $\mathbf{x}$; router weights $\mathbf{W}_r$; \#groups $n$; experts/group $m$; active $k$
\STATE {\bfseries Output:} selected experts $\mathcal{S}$, weights $\mathbf{w}$
\STATE $\mathbf{s} \leftarrow \mathbf{W}_r \mathbf{x}$ \hfill $\triangleright$ Router scores $\in \mathbb{R}^{n \times m}$
\STATE \textit{// Stage 1: Intra-group selection}
\FOR{$i = 1$ {\bfseries to} $n$}
    \STATE $j_i^* \leftarrow \operatorname{argmax}_{j \in [m]} s_{i,j}$ \hfill $\triangleright$ Best expert in group $i$
    \STATE $s_i^* \leftarrow s_{i, j_i^*}$ \hfill $\triangleright$ Representative score
\ENDFOR
\STATE \textit{// Stage 2: Inter-group selection}
\STATE $\mathcal{G} \leftarrow \text{TopK}(\{s_1^*, \ldots, s_n^*\}, k)$ \hfill $\triangleright$ Top-$k$ groups
\STATE $\mathcal{S} \leftarrow \{(i, j_i^*) : i \in \mathcal{G}\}$ \hfill $\triangleright$ Selected experts
\STATE \textit{// Compute normalized routing weights}
\FOR{$i \in \mathcal{G}$}
    \STATE $w_i \leftarrow \exp(s_i^*) / \sum_{j \in \mathcal{G}} \exp(s_j^*)$
\ENDFOR
\STATE {\bfseries Return} $\mathcal{S}$, $\mathbf{w}$
\end{algorithmic}
\end{algorithm}

Selected expert outputs are aggregated with normalized weights $\mathbf{y} = \sum_{i \in \mathcal{S}} w_i \cdot \text{Expert}_{i,j_i^*}(\mathbf{x})$. The auxiliary load-balancing loss takes the standard form $\mathcal{L}_{\text{aux}} = \alpha \cdot N \cdot \sum_{i=1}^{N} f_i \cdot p_i$~\citep{fedus2022switch}, where $f_i$ is the fraction of tokens routed to expert $i$ and $p_i$ its mean routing probability.

\subsection{Training Pipeline}
\label{sec:training}

The \divmoe{} pipeline has three stages, all using only public corpora. \emph{Stage~1} (domain-specialist initialization) continually pre-trains four copies of the dense base on the per-domain partitions of Dolmino-Mix~\citep{olmo2025dolmino}---math, code, academic, web---for $4{\times}10\text{B}{=}40\text{B}$ tokens, producing the specialists from which fine-grained experts are sliced. \emph{Stage~2} (joint continual pre-training of the assembled MoE) trains the full \divmoe{} model on \texttt{dolmino-mix}~\citep{olmo2025dolmino,soldaini2024dolma} for 500B tokens; the corpus, optimizer, and step count are identical for every upcycling baseline, ensuring a controlled comparison. \emph{Stage~3} (supervised fine-tuning, used \emph{only} for the open-source MoE comparison in \Cref{sec:moe_comparison}) applies T\"ulu-3-SFT-Mixture~\citep{lambert2024tulu3} identically to \divmoe{}, NVIDIA-Upcycling, and OLMoE, isolating architectural effects from data effects. Per-stage corpus, hyperparameter, and compute details are provided in \Cref{app:pipeline,app:hyperparams,app:compute}.

\section{Experiments}
\label{sec:experiments}
\subsection{Experimental Setup}
\label{sec:exp_setup}
\paragraph{Models and architecture.}
We evaluate two base models, Qwen3-1.7B-Base~\citep{qwen2024qwen3} and Llama3.2-1B, and additionally upcycle Qwen3-4B into a 12B-parameter \divmoe{} for the open-source MoE comparison. All variants use $n{=}4$ domain groups, $m{=}2$ experts per group ($N{=}8$ total), and $k{=}2$ active experts per token.

\paragraph{Baselines.}
We compare against six MoE construction methods, all sharing the Stage~2 CPT corpus, optimizer, and step count: \textbf{From Scratch (FS)} (random fine-grained MoE init), \textbf{Sparse Upcycling (SU)}~\citep{komatsuzaki2022sparse}, \textbf{Branch-Train-Mix (BTX)}~\citep{sukhbaatar2024branch} (coarse-grained MoE assembled from domain branch models), \textbf{NVIDIA Upcycling (NU)}~\citep{he2024upcycling} (fine-grained, dense-equivalent, single-source), \textbf{Drop-Upcycling (DU)}~\citep{chen2024dropupcycling} (coarse, with random re-initialization), and its fine-grained variant \textbf{DU-f}.

\paragraph{Evaluation.}
We evaluate on 15 benchmarks spanning general knowledge, reasoning, code, and graduate-level science (ARC-c/e~\citep{clark2018think}, BoolQ~\citep{clark2019boolq}, COPA~\citep{roemmele2011choice}, MMLU~\citep{hendrycks2021measuring}, OpenBookQA~\citep{mihaylov2018can}, TriviaQA~\citep{joshi2017triviaqa}, HellaSwag~\citep{zellers2019hellaswag}, SQuAD2~\citep{rajpurkar2018know}, GSM8K~\citep{cobbe2021training}, MATH~\citep{hendrycks2021measuringmathematicalproblemsolving}, BBH~\citep{suzgun2022bbh}, HumanEval~\citep{chen2021evaluating}, MBPP~\citep{austin2021program}, GPQA~\citep{rein2024gpqa}). Training hyperparameters, per-method compute, and per-benchmark shot counts are tabulated in \Cref{app:hyperparams,app:compute,app:eval_config}.
\subsection{Comparison with Upcycling Methods}
\label{sec:upcycling_comparison}

\Cref{tab:upcycling_qwen,tab:upcycling_llama} report results after Stage~2 continual pre-training (no SFT) on a representative subset of 11 benchmarks. All upcycling methods share the same Stage~2 corpus, optimizer, and step count; the only differences are the initialization and routing schemes. More results are provided in \Cref{app:full_upcycling}.

\begin{table}[t]
\caption{Upcycling comparison on Qwen3-1.7B-Base after 500B Stage~2 CPT tokens (540B for BTX and \divmoe{}, including Stage~1). Best across upcycling methods in \textbf{bold}. F.G.\ denotes fine-grained partitioning. ``Avg.''\ averages over the 11 benchmarks shown; full 15-benchmark results in \Cref{app:full_upcycling}.}
\label{tab:upcycling_qwen}
\centering
\scriptsize
\setlength{\tabcolsep}{2.6pt}
\begin{tabular}{lc|ccccc|cccc|cc|c}
\toprule
\textbf{Method} & \textbf{F.G.} & \textbf{ARC-c} & \textbf{BoolQ} & \textbf{MMLU} & \textbf{TriviaQ} & \textbf{HellaS.} & \textbf{GSM8K} & \textbf{MATH} & \textbf{HumanE.} & \textbf{MBPP} & \textbf{GPQA} & \textbf{BBH} & \textbf{Avg.} \\
\midrule
From Scratch                   & \cmark  & 12.2 & 30.5 & 40.5 & 33.0 & 42.5 & 5.5  & 2.3  & 6.8  & 11.2 & 22.6 & 20.5 & 20.7 \\
Sparse Upcycling               & \xmark  & 38.5 & 60.2 & 56.8 & 43.5 & 54.5 & 56.5 & 28.5 & 38.0 & 47.5 & 24.5 & 47.5 & 45.1 \\
BTX                            & \xmark  & 43.0 & 71.0 & 61.5 & 47.5 & 61.0 & 64.0 & 36.5 & 44.0 & 53.0 & 26.0 & 51.5 & 50.8 \\
NVIDIA Upcycling               & \cmark  & 42.5 & 70.5 & 61.5 & 47.5 & 60.5 & 62.5 & 35.5 & 44.5 & 53.5 & 25.8 & 51.0 & 50.5 \\
Drop-Upcycling                 & \xmark  & 41.5 & 71.5 & 61.0 & 47.0 & 59.5 & 60.0 & 33.5 & 41.0 & 50.5 & 25.0 & 50.0 & 49.1 \\
Drop-Upcycling-f               & \cmark  & 14.5 & 32.5 & 42.0 & 33.5 & 41.5 & 7.0  & 3.5  & 7.8  & 12.5 & 22.5 & 22.0 & 21.8 \\
\rowcolor{gray!15}
\textbf{\divmoe{}}             & \cmark & \textbf{45.0} & \textbf{75.0} & \textbf{64.5} & \textbf{50.0} & \textbf{63.5} & \textbf{68.5} & \textbf{40.5} & \textbf{49.0} & \textbf{57.5} & \textbf{28.0} & \textbf{55.5} & \textbf{54.3} \\
\bottomrule
\end{tabular}
\end{table}

\begin{table}[t]
\caption{Upcycling comparison on Llama3.2-1B after 500B Stage~2 CPT tokens (540B for BTX and \divmoe{}). Same conventions as \Cref{tab:upcycling_qwen}.}
\label{tab:upcycling_llama}
\centering
\scriptsize
\setlength{\tabcolsep}{2.6pt}
\begin{tabular}{lc|ccccc|cccc|cc|c}
\toprule
\textbf{Method} & \textbf{F.G.} & \textbf{ARC-c} & \textbf{BoolQ} & \textbf{MMLU} & \textbf{TriviaQ} & \textbf{HellaS.} & \textbf{GSM8K} & \textbf{MATH} & \textbf{HumanE.} & \textbf{MBPP} & \textbf{GPQA} & \textbf{BBH} & \textbf{Avg.} \\
\midrule
From Scratch                   & \cmark  & 13.0 & 27.5 & 38.0 & 31.0 & 36.0 & 4.0  & 1.5  & 4.5  & 8.0  & 21.8 & 18.0 & 18.5 \\
Sparse Upcycling               & \xmark  & 33.5 & 60.5 & 48.0 & 39.5 & 52.5 & 26.5 & 9.0  & 15.0 & 21.5 & 23.5 & 31.5 & 32.8 \\
BTX                            & \xmark  & 36.0 & 66.0 & 50.5 & 39.5 & 56.0 & 31.5 & 11.0 & 17.5 & 24.5 & 25.0 & 33.0 & 35.5 \\
NVIDIA Upcycling               & \cmark  & 36.0 & 65.5 & 50.0 & 39.0 & 55.5 & 30.5 & 10.8 & 17.0 & 24.0 & 24.8 & 32.8 & 35.1 \\
Drop-Upcycling                 & \xmark  & 35.0 & 66.0 & 49.5 & 39.0 & 55.0 & 29.5 & 10.5 & 16.5 & 23.5 & 24.5 & 32.5 & 34.7 \\
Drop-Upcycling-f               & \cmark  & 25.5 & 33.0 & 43.5 & 38.0 & 39.5 & 6.0  & 2.5  & 6.0  & 9.5  & 21.5 & 19.0 & 22.2 \\
\rowcolor{gray!15}
\textbf{\divmoe{}}             & \cmark & \textbf{37.5} & \textbf{67.5} & \textbf{52.5} & \textbf{42.0} & \textbf{56.5} & \textbf{34.5} & \textbf{13.5} & \textbf{20.0} & \textbf{27.0} & \textbf{26.0} & \textbf{34.5} & \textbf{37.4} \\
\bottomrule
\end{tabular}
\end{table}

\paragraph{\divmoe{} achieves the best overall performance.}
On Qwen3-1.7B-Base, \divmoe{} reaches 54.3\% averaged over the 11 main-paper benchmarks, +3.5 points over the strongest baseline (BTX, 50.8\%) and +3.8 over NVIDIA Upcycling (50.5\%). The pattern holds on Llama3.2-1B (37.4\% vs.\ 35.5\% for BTX). The advantage is largest on reasoning and code: +4.5 points on GSM8K, +4.0 on MATH, and +5.0 on HumanEval over BTX (Qwen3-1.7B-Base), confirming that domain-specialized initialization transfers reasoning capability into the assembled MoE. The gap is sustained throughout training rather than driven by initialization alone (\Cref{app:learning_main}).

\paragraph{Stage~2 CPT improves over the dense base.}
Unlike prior work where post-CPT upcycled models regress relative to the dense base on most benchmarks~\citep{komatsuzaki2022sparse,he2024upcycling,chen2024dropupcycling}, our use of \texttt{dolmino-mix} as the CPT corpus eliminates this regression: \divmoe{} surpasses Qwen3-1.7B-Base on every benchmark we evaluate (e.g., MMLU 64.5 vs.\ 62.6, MATH 40.5 vs.\ 38.4, BBH 55.5 vs.\ 53.5; full reference rows in \Cref{app:full_upcycling}). \divmoe{} also outperforms a data-matched dense baseline (continually pre-trained on the same 540B tokens, no MoE conversion) by $\sim$2 points despite activating only $\sim$2.4B of its 5.4B parameters per token, demonstrating that the MoE architecture provides value beyond the additional CPT data.

\paragraph{Fine-grained collapse without domain diversity.}
From-Scratch (20.7\%) and Drop-Upcycling-fine-grained (21.8\%) collapse to near-random performance on Qwen3-1.7B-Base---essentially identical, despite very different initialization schemes (random vs.\ partial re-initialization). The pattern repeats on Llama3.2-1B (18.5\% and 22.2\%). Coarse-grained Drop-Upcycling (49.1\%) does \emph{not} collapse, isolating the failure to the combination of fine-grained partitioning with non-domain-specialized initialization. \divmoe{}'s domain-specialized initialization is the only mechanism in our experiments that recovers fine-grained capacity.

\subsection{Comparison with Open-Source MoE Models (Controlled SFT)}
\label{sec:moe_comparison}

We next compare against state-of-the-art open-source MoE checkpoints: OLMoE~\citep{muennighoff2024olmoe}, DeepSeek-V2-Lite~\citep{liu2024deepseekv2}, and Moonlight-MoE~\citep{liu2025moonlight}. A direct comparison would be confounded by the difference in fine-tuning data: the public checkpoints use their own SFT mixtures, while our model is fine-tuned on T\"ulu-3-SFT-Mixture. To isolate the architectural contribution from the data effect, we additionally fine-tune NVIDIA Upcycling (the strongest fine-grained baseline) and OLMoE on the \emph{same} T\"ulu-3-SFT-Mixture under identical Stage~3 hyperparameters as \divmoe{} and report the controlled rows in \Cref{tab:opensource_moe}.

\begin{table}[t]
\caption{Comparison with open-source MoE models on a representative 7-benchmark subset. Size denotes active/total parameters. \divmoe{} is upcycled from Qwen3-4B-Base; all baselines use released checkpoints. Rows marked ``+T\"ulu-3 SFT'' apply the \emph{same} fine-tuning to a baseline. Best in \textbf{bold}, second-best \underline{underlined}. Full 8-benchmark results (including ARC-e) in \Cref{app:opensource_moe_full}.}
\label{tab:opensource_moe}
\centering
\scriptsize
\setlength{\tabcolsep}{2.6pt}
\begin{tabular}{lc|ccc|cccc|c}
\toprule
\textbf{Model} & \textbf{Size} & \textbf{ARC-c} & \textbf{MMLU} & \textbf{TriviaQA} & \textbf{HumanE.} & \textbf{MBPP} & \textbf{GSM8K} & \textbf{MATH} & \textbf{Avg.} \\
\midrule
OLMoE                                  & 1B/7B    & 49.2          & 51.9          & 60.4          & \textbf{51.8} & 61.2          & 45.5          & 23.9          & 49.1 \\
OLMoE + T\"ulu-3 SFT                   & 1B/7B    & 52.4          & 55.8          & 63.5          & \underline{53.5} & 62.8          & 56.8          & 31.2          & 53.7 \\
DeepSeek-V2-Lite                       & 2.4B/16B & 52.1          & 58.3          & 65.1          & 29.9          & 43.2          & 41.1          & 17.1          & 43.8 \\
NVIDIA Up. + T\"ulu-3 SFT              & 2.4B/5.4B& 56.3          & 64.2          & 67.5          & 45.8          & 54.2          & 66.5          & 38.8          & 56.2 \\
Moonlight-MoE                          & 2.4B/16B & \textbf{65.5} & \underline{70.0} & \underline{66.3} & 48.1 & \textbf{63.8} & \textbf{77.4} & \underline{45.3} & \underline{62.3} \\
\rowcolor{gray!15}
\textbf{\divmoe{}}                     & 3.6B/12B & \underline{63.1} & \textbf{71.5} & \textbf{73.3} & \underline{51.4} & \underline{60.0} & \underline{74.4} & \textbf{47.5} & \textbf{63.0} \\
\bottomrule
\end{tabular}
\end{table}

\paragraph{Architecture-isolated advantage over the controlled baseline.}
\Cref{tab:opensource_moe} shows that with the same Stage~3 SFT, \divmoe{} outperforms NVIDIA Upcycling by +6.8 points (63.0 vs.\ 56.2) and OLMoE by +9.3 points (63.0 vs.\ 53.7). The largest controlled gaps are on GSM8K (+7.9 over NVIDIA Up.) and MATH (+8.7), indicating that the architectural advantage is concentrated on multi-step reasoning where cross-domain composition is most useful.

\paragraph{Match against Moonlight-MoE despite smaller size.}
\divmoe{} matches Moonlight-MoE (16B total) using only 12B total parameters (63.0 vs.\ 62.3 on the 7-benchmark subset; 64.5 vs.\ 64.5 on the full 8-benchmark set, \Cref{app:opensource_moe_full}), while leading on MMLU (+1.5), TriviaQA (+7.0), and MATH (+2.2). Moonlight-MoE was trained from scratch with substantially more compute; \divmoe{}'s upcycling pipeline reaches the same overall performance with only 540B Stage-2 CPT tokens.

\section{Ablation Studies}
\label{sec:ablation}

\subsection{Component Synergy and the Diversity Constraint}
\label{sec:ablation_synergy}\label{sec:ablation_diversity}

We isolate each design decision against the full \divmoe{} on Qwen3-1.7B-Base (\Cref{tab:synergy}). Removing \emph{any} of the three design axes reduces performance, and removing two collapses fine-grained variants to near-random; the combination is far from a trivial union, solving a failure mode (catastrophic collapse) that no individual component addresses. We further ask whether forcing cross-domain co-activation \emph{harms} domain-specific tasks---a natural concern given that, e.g., a coding token is now obliged to also draw from math, science, or commonsense experts. \Cref{tab:per_domain} shows the opposite: the constraint \emph{improves} performance on every domain-specific benchmark evaluated.

\begin{table}[t]
\centering
\begin{minipage}[t]{0.54\linewidth}
\centering
\caption{Component synergy ablation on Qwen3-1.7B-Base. ``$-$ DC'' removes the diversity constraint; ``DU-f'' removes domain diversity; ``BTX'' removes fine-grained partitioning.}
\label{tab:synergy}
\scriptsize
\setlength{\tabcolsep}{2.6pt}
\begin{tabular}{lccc|c}
\toprule
\textbf{Configuration} & \textbf{Domain} & \textbf{Fine-} & \textbf{Diversity} & \textbf{Avg.} \\
                        & \textbf{experts} & \textbf{grained} & \textbf{constraint} & \textbf{(15 bm.)} \\
\midrule
Drop-Upcycling-f      & \xmark & \cmark & \xmark & 23.2 \\
NVIDIA Upcycling      & \xmark & \cmark & \xmark & 51.1 \\
BTX                   & \cmark & \xmark & \xmark & 51.6 \\
\divmoe{} ($-$ DC)    & \cmark & \cmark & \xmark & 52.5 \\
\rowcolor{gray!15}
\textbf{\divmoe{}}    & \cmark & \cmark & \cmark & \textbf{55.6} \\
\bottomrule
\end{tabular}
\end{minipage}\hfill
\begin{minipage}[t]{0.44\linewidth}
\centering
\caption{Effect of the diversity constraint on domain-specific benchmarks (Qwen3-1.7B-Base). The constraint improves all four metrics.}
\label{tab:per_domain}
\scriptsize
\setlength{\tabcolsep}{2.6pt}
\begin{tabular}{l|cccc}
\toprule
\textbf{Method} & \textbf{HumanE.} & \textbf{MBPP} & \textbf{GSM8K} & \textbf{MATH} \\
\midrule
\divmoe{} w/o DC & 23.8 & 31.5 & 43.2 & 18.1 \\
\rowcolor{gray!15}
\divmoe{} w/ DC & \textbf{25.6} & \textbf{33.8} & \textbf{45.3} & \textbf{19.5} \\
$\Delta$        & +1.8 & +2.3 & +2.1 & +1.4 \\
\bottomrule
\end{tabular}
\end{minipage}
\end{table}

\subsection{Routing Patterns}
\label{sec:routing_patterns}

\begin{figure}[t]
\centering
\includegraphics[width=0.99\linewidth]{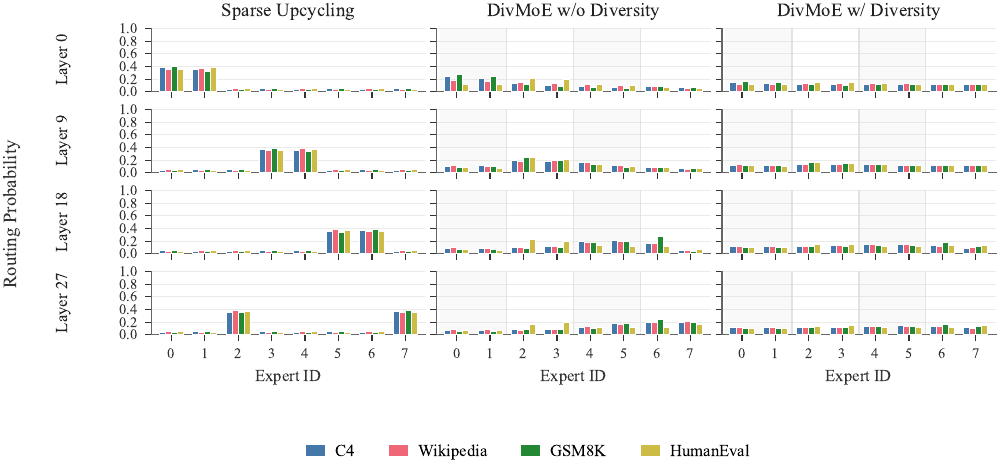}
\caption{Expert routing patterns across layers and datasets. Sparse Upcycling exhibits routing collapse with few experts dominating. \divmoe{} without the constraint already shows improved utilization. With the constraint, routing is balanced across domain groups preserving dataset-specific preferences: GSM8K activates more math experts, while HumanEval activates more code experts.}
\label{fig:routing_patterns}
\end{figure}

\Cref{fig:routing_patterns} visualizes routing probabilities across experts for Sparse Upcycling, \divmoe{} without the constraint, and full \divmoe{}. Sparse Upcycling shows routing collapse in deeper layers (over 60\% of tokens routed to two experts at layer 36). \divmoe{} without the constraint already shows improved utilization due to domain-specialized initialization. With the constraint, routing is balanced \emph{across} groups while preserving \emph{within}-group specialization (\Cref{sec:ablation_diversity}).

\subsection{Effect of Expert Granularity}
\label{sec:ablation_granularity}
\begin{figure}[t]
\centering
\includegraphics[width=0.47\linewidth]{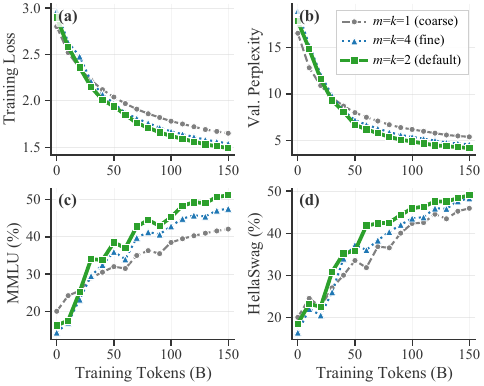}
\caption{Effect of expert granularity ($m=k$) on Qwen3-1.7B-Base. Medium granularity ($m=k=2$) provides the best balance between training-loss reduction and downstream accuracy.}
\label{fig:granularity}
\end{figure}

We vary $m=k\in\{1,2,4\}$ (\Cref{fig:granularity}). Increasing granularity from 1 to 4 reduces training loss progressively, but $m=k=4$ shows mild overfitting on MMLU/HellaSwag. We adopt $m=k=2$ as the default, which balances capacity with generalization while keeping the active-parameter footprint at half of the dense base.

\section{Conclusion}
\label{sec:conclusion}

We presented \divmoe{}, the first framework for fine-grained MoE upcycling that addresses the previously-unobserved \emph{fine-grained upcycling collapse} pathology: combining fine-grained expert partitioning with single-source initialization causes downstream accuracy to drop to near-random, irrespective of the partitioning scheme. \divmoe{} resolves this issue through three components: (i)~domain-specialized fine-grained expert initialization, (ii)~a hard diversity-constrained routing mechanism, and (iii)~a public-corpus continual pre-training pipeline that no longer regresses below the dense base. Ablation studies demonstrate that the proposed components are complementary and jointly necessary for robust fine-grained upcycling. Across comprehensive experiments, \divmoe{} consistently achieves the strongest overall performance. Notably, the structural constraint improves performance on every domain-specific benchmark we test, refuting the intuition that cross-domain co-activation should harm specialized tasks.

\begin{ack}
Yang You’s research group is being sponsored by NUS startup grant (Presidential Young Professorship), Singapore MOE Tier-1 grant, ByteDance grant, ARCTIC grant, SMI grant and Alibaba grant.
This work is also supported by NSF 2048280, 2325121, 2244760, 2331966 and ONR N0001423-1-2300:P00001. Additionally, We thank the open-source community for releasing code and datasets that made this research possible.
\end{ack}


\newpage
\appendix

\section{Limitations and Future Work}
\label{app:limitations}
Our experiments use base models up to 4B parameters, yielding MoE models up to 12B; scaling to 7B-13B base models is an important direction. Our domain partition is fixed to four broad domains; learning the partition end-to-end is left to future work. Finally, \divmoe{} requires an additional 40B Stage-1 token budget per base model, an 8\% overhead relative to from-scratch training, which is small in absolute terms but non-trivial for very-large-scale settings.
\section{Implementation Details}
\label{app:implementation}

\subsection{Model Architecture}

\Cref{tab:model_arch} summarizes the architecture configurations for each base model. All \divmoe{} variants use $n=4$ domain groups, $m=2$ experts per group ($N=8$ total), and $k=2$ active experts per token.

\begin{table}[h]
\caption{Model architecture configurations. \divmoe{} parameters are derived from each base model with expansion factor $n=4$.}
\label{tab:model_arch}
\centering
\small
\begin{tabular}{lccc}
\toprule
\textbf{Configuration} & \textbf{Qwen3-4B} & \textbf{Qwen3-1.7B} & \textbf{Llama3.2-1B} \\
\midrule
Hidden size & 2560 & 2048 & 2048 \\
FFN intermediate size & 6912 & 5632 & 8192 \\
Num layers & 36 & 28 & 16 \\
Num attention heads & 32 & 16 & 32 \\
Num KV heads & 4 & 4 & 8 \\
Vocab size & 151{,}936 & 151{,}936 & 128{,}256 \\
\midrule
\multicolumn{4}{c}{\textit{\divmoe{} Configuration}} \\
\midrule
Domain groups ($n$)        & 4 & 4 & 4 \\
Experts per group ($m$)    & 2 & 2 & 2 \\
Total experts ($N$)        & 8 & 8 & 8 \\
Active experts ($k$)       & 2 & 2 & 2 \\
Expert FFN dim             & 3456 & 2816 & 4096 \\
Dense params               & 4B   & 1.7B & 1B   \\
Total MoE params           & 12B  & 5.4B & 3.2B \\
Active params              & 3.6B & 2.4B & 1.8B \\
\bottomrule
\end{tabular}
\end{table}

\subsection{Training Pipeline Details}
\label{app:pipeline}

This section expands \Cref{sec:training} with full descriptions of each pipeline stage. Throughout, we use only \emph{public corpora} so all results are independently reproducible.

\paragraph{Stage 1: Domain-specialist initialization (40B tokens).}
We continually pre-train four copies of the base model (next-token-prediction loss on full sequences, not instruction-style SFT) on domain-specific subsets of public pre-training corpora, producing $\mathcal{M}_{\text{math}}$, $\mathcal{M}_{\text{code}}$, $\mathcal{M}_{\text{science}}$, $\mathcal{M}_{\text{common}}$. We use the per-domain partitions of Dolmino-Mix~\citep{olmo2025dolmino}---specifically its \texttt{math}, \texttt{code}, \texttt{academic}, and \texttt{web} subsets---which jointly cover the four target domains. Each specialist sees 10B tokens (40B total). This is the only stage where the base model weights are updated as a dense model; subsequent stages train the assembled MoE.

\paragraph{Stage 2: Continual pre-training of the assembled MoE (500B tokens).}
After constructing the \divmoe{} architecture (\Cref{sec:initialization}), we continually pre-train the entire MoE on \texttt{allenai/dolma3\_dolmino\_mix-100B-1025}~\citep{olmo2025dolmino,soldaini2024dolma}, a high-quality public mid-training corpus, for 500B tokens. This stage adapts the model to sparse computation, learns the router weights, and enables cross-expert collaboration. Crucially, this is the same corpus, hyperparameters, and step count used for every upcycling baseline, ensuring controlled comparison.

\paragraph{Stage 3: Supervised fine-tuning (used only for the open-source MoE comparison).}
For the comparison against open-source MoE checkpoints in \Cref{sec:moe_comparison}, we additionally fine-tune \divmoe{} on T\"ulu-3-SFT-Mixture~\citep{lambert2024tulu3}, a public reasoning-and-instruction mixture, with sequence length 16{,}384. To isolate architectural effects from data effects, we apply the \emph{same} Stage~3 SFT to the strongest baseline (NVIDIA Upcycling) and to OLMoE; the controlled comparison is reported in \Cref{tab:opensource_moe}.

\paragraph{No private data.}
The pipeline uses only the dense base checkpoints, Dolmino-Mix-1124, and T\"ulu-3-SFT-Mixture, all of which are publicly available; we release the slicing and routing code together with the per-domain Dolmino-Mix splits.

\subsection{Training Hyperparameters}
\label{app:hyperparams}

\Cref{tab:hyperparams} details the hyperparameters for each stage. All experiments use AdamW ($\beta_1=0.9$, $\beta_2=0.95$) with bf16 mixed precision and global batch size of 2M tokens.

\begin{table}[h]
\caption{Training hyperparameters by stage.}
\label{tab:hyperparams}
\centering
\small
\begin{tabular}{lccc}
\toprule
\textbf{Hyperparameter} & \textbf{Stage 1} & \textbf{Stage 2} & \textbf{Stage 3} \\
                         & \textit{Domain CPT} & \textit{Joint CPT} & \textit{SFT} \\
\midrule
Learning rate            & $5{\times}10^{-5}$ & $5{\times}10^{-4}$ & $2{\times}10^{-5}$ \\
Min learning rate        & $5{\times}10^{-6}$ & $5{\times}10^{-5}$ & $2{\times}10^{-6}$ \\
LR scheduler             & Cosine & Cosine & Cosine \\
Warmup steps             & 50     & 256    & 500 \\
Weight decay             & 0.1    & 0.1    & 0.1 \\
Sequence length          & 4{,}096 & 4{,}096 & 16{,}384 \\
Training tokens          & 10B per domain (40B total) & 500B & -- \\
Gradient checkpointing   & \cmark & \cmark & \cmark \\
\bottomrule
\end{tabular}
\end{table}

For Stage~3 with long sequences, we use DeepSpeed Ulysses~\citep{jacobs2023deepspeedulyssesoptimizationsenabling} sequence parallelism (factor 4) and CPU offloading of optimizer states.

\subsection{Compute Budget}
\label{app:compute}

\Cref{tab:compute} compares compute budgets across methods. \divmoe{} and BTX share identical total compute (540B tokens; the additional 40B Stage-1 tokens are dense-model compute, substantially cheaper per-token than MoE training).

\begin{table}[h]
\caption{Per-method compute. GPU-hours measured on NVIDIA H20 (80GB), Qwen3-1.7B-Base.}
\label{tab:compute}
\centering
\small
\begin{tabular}{lcccc}
\toprule
\textbf{Method} & \textbf{Stage-1 Tokens} & \textbf{Stage-2 Tokens} & \textbf{Total} & \textbf{GPU-hours} \\
\midrule
From Scratch        & 0          & 500B (MoE) & 500B & $\sim$26{,}500 \\
Sparse Upcycling    & 0          & 500B (MoE) & 500B & $\sim$26{,}500 \\
BTX                 & 40B (dense) & 500B (MoE) & 540B & $\sim$28{,}800 \\
NVIDIA Upcycling    & 0          & 500B (MoE) & 500B & $\sim$26{,}500 \\
Drop-Upcycling      & 0          & 500B (MoE) & 500B & $\sim$26{,}500 \\
\divmoe{}           & 40B (dense) & 500B (MoE) & 540B & $\sim$28{,}800 \\
\bottomrule
\end{tabular}
\end{table}

\subsection{Evaluation Configuration}
\label{app:eval_config}

We evaluate using LM Evaluation Harness~\citep{eval-harness} with greedy decoding (temperature 0) for generative tasks and likelihood-based scoring for multiple-choice tasks. \Cref{tab:eval_config} lists the format and shot count for each benchmark.

\begin{table}[h]
\caption{Evaluation format and shot count per benchmark.}
\label{tab:eval_config}
\centering
\small
\begin{tabular}{lcc}
\toprule
\textbf{Benchmark} & \textbf{Format} & \textbf{Shots} \\
\midrule
ARC-Challenge & Likelihood & 25 \\
ARC-Easy      & Likelihood & 25 \\
BoolQ         & Likelihood & 0 \\
COPA          & Likelihood & 0 \\
MMLU          & Likelihood & 5 \\
OpenBookQA    & Likelihood & 0 \\
TriviaQA      & Generative & 5 \\
HellaSwag     & Likelihood & 10 \\
SQuAD2        & Generative & 1 \\
HumanEval     & Generative & 0 \\
MBPP          & Generative & 3 \\
GSM8K         & Generative & 8 \\
MATH          & Generative & 4 \\
GPQA          & Likelihood & 0 \\
BBH           & Generative & 3 \\
\bottomrule
\end{tabular}
\end{table}

\section{Full Upcycling Results (15 Benchmarks)}
\label{app:full_upcycling}

Tables~\ref{tab:upcycling_qwen_full} and \ref{tab:upcycling_llama_full} report the complete 15-benchmark upcycling comparison underlying the abridged \Cref{tab:upcycling_qwen,tab:upcycling_llama} in the main paper. We additionally include reference rows for the dense base model, the four individual domain specialists, and a \emph{data-matched dense baseline} (the dense base continually pre-trained on the same 540B tokens, no MoE conversion). These rows isolate the effect of the MoE architecture from the effect of the additional CPT data.

\begin{table}[h]
\caption{Full upcycling comparison on Qwen3-1.7B-Base after 500B Stage~2 CPT tokens (540B for BTX and \divmoe{}, including Stage~1). Best across upcycling methods in \textbf{bold}; reference rows (above the rule) provide context. F.G.\ denotes fine-grained partitioning.}
\label{tab:upcycling_qwen_full}
\centering
\scriptsize
\setlength{\tabcolsep}{2.6pt}
\resizebox{\textwidth}{!}{
\begin{tabular}{lc|ccccccccc|cccccc|c}
\toprule
\textbf{Method} & \textbf{F.G.} & \textbf{ARC-c} & \textbf{ARC-e} & \textbf{BoolQ} & \textbf{COPA} & \textbf{MMLU} & \textbf{OBQA} & \textbf{TriviaQ} & \textbf{HellaS.} & \textbf{SQuAD2} & \textbf{GSM8K} & \textbf{MATH} & \textbf{HumanE.} & \textbf{MBPP} & \textbf{GPQA} & \textbf{BBH} & \textbf{Avg.} \\
\midrule
\multicolumn{18}{l}{\textit{Reference rows (not upcycling baselines)}} \\
Qwen3-1.7B-Base                & --      & 43.2 & 72.4 & 72.8 & 76.0 & 62.6 & 38.6 & 48.5 & 63.2 & 32.5 & 65.8 & 38.4 & 46.5 & 54.8 & 26.5 & 53.5 & 53.0 \\
Math specialist                & --      & 39.8 & 60.5 & 64.8 & 68.2 & 59.5 & 34.2 & 37.8 & 56.5 & 25.8 & 76.2 & 48.5 & 40.3 & 47.5 & 27.8 & 49.6 & 49.1 \\
Code specialist                & --      & 36.5 & 55.2 & 60.4 & 64.8 & 54.1 & 32.5 & 34.8 & 53.8 & 23.5 & 52.4 & 30.5 & 57.8 & 61.5 & 24.2 & 44.3 & 45.8 \\
Science specialist             & --      & 46.8 & 73.2 & 70.5 & 73.8 & 61.8 & 42.5 & 44.2 & 61.5 & 30.2 & 60.5 & 37.2 & 39.5 & 48.3 & 31.8 & 52.6 & 51.6 \\
Commonsense specialist         & --      & 42.2 & 68.5 & 76.2 & 80.5 & 59.2 & 40.8 & 46.5 & 66.2 & 33.5 & 55.3 & 33.8 & 37.2 & 46.1 & 25.0 & 54.8 & 51.1 \\
Dense baseline (same data)     & --      & 44.0 & 73.1 & 73.4 & 76.3 & 63.2 & 39.0 & 48.8 & 63.0 & 32.7 & 66.0 & 38.7 & 47.0 & 55.4 & 26.8 & 53.8 & 53.4 \\
\midrule
\multicolumn{18}{l}{\textit{Upcycling methods}} \\
From Scratch                   & \cmark  & 12.2 & 22.0 & 30.5 & 38.7 & 40.5 & 27.0 & 33.0 & 42.5 & 17.2 & 5.5  & 2.3  & 6.8  & 11.2 & 22.6 & 20.5 & 22.2 \\
Sparse Upcycling               & \xmark  & 38.5 & 62.8 & 60.2 & 65.4 & 56.8 & 35.0 & 43.5 & 54.5 & 27.0 & 56.5 & 28.5 & 38.0 & 47.5 & 24.5 & 47.5 & 45.7 \\
BTX                            & \xmark  & 43.0 & 70.5 & 71.0 & 74.0 & 61.5 & 39.5 & 47.5 & 61.0 & 30.5 & 64.0 & 36.5 & 44.0 & 53.0 & 26.0 & 51.5 & 51.6 \\
NVIDIA Upcycling               & \cmark  & 42.5 & 69.5 & 70.5 & 73.0 & 61.5 & 38.5 & 47.5 & 60.5 & 30.0 & 62.5 & 35.5 & 44.5 & 53.5 & 25.8 & 51.0 & 51.1 \\
Drop-Upcycling                 & \xmark  & 41.5 & 69.0 & 71.5 & 75.0 & 61.0 & 37.0 & 47.0 & 59.5 & 31.5 & 60.0 & 33.5 & 41.0 & 50.5 & 25.0 & 50.0 & 50.2 \\
Drop-Upcycling-f               & \cmark  & 14.5 & 24.5 & 32.5 & 41.5 & 42.0 & 26.0 & 33.5 & 41.5 & 17.5 & 7.0  & 3.5  & 7.8  & 12.5 & 22.5 & 22.0 & 23.2 \\
\rowcolor{gray!15}
\textbf{\divmoe{}}             & \cmark & \textbf{45.0} & \textbf{73.5} & \textbf{75.0} & \textbf{78.0} & \textbf{64.5} & \textbf{42.0} & \textbf{50.0} & \textbf{63.5} & \textbf{33.5} & \textbf{68.5} & \textbf{40.5} & \textbf{49.0} & \textbf{57.5} & \textbf{28.0} & \textbf{55.5} & \textbf{55.6} \\
\bottomrule
\end{tabular}
}
\end{table}

\begin{table}[h]
\caption{Full upcycling comparison on Llama3.2-1B after 500B Stage~2 CPT tokens (540B for BTX and \divmoe{}). Same conventions as \Cref{tab:upcycling_qwen_full}.}
\label{tab:upcycling_llama_full}
\centering
\scriptsize
\setlength{\tabcolsep}{2.6pt}
\resizebox{\textwidth}{!}{
\begin{tabular}{lc|ccccccccc|cccccc|c}
\toprule
\textbf{Method} & \textbf{F.G.} & \textbf{ARC-c} & \textbf{ARC-e} & \textbf{BoolQ} & \textbf{COPA} & \textbf{MMLU} & \textbf{OBQA} & \textbf{TriviaQ} & \textbf{HellaS.} & \textbf{SQuAD2} & \textbf{GSM8K} & \textbf{MATH} & \textbf{HumanE.} & \textbf{MBPP} & \textbf{GPQA} & \textbf{BBH} & \textbf{Avg.} \\
\midrule
Llama3.2-1B (ref.)             & --      & 34.8 & 63.5 & 65.2 & 68.0 & 49.3 & 32.4 & 38.5 & 55.2 & 28.5 & 28.5 & 10.2 & 16.5 & 23.8 & 24.8 & 32.4 & 38.1 \\
Dense baseline (same data)     & --      & 35.2 & 63.8 & 65.7 & 68.5 & 50.0 & 32.8 & 39.0 & 55.4 & 28.8 & 29.0 & 10.5 & 17.0 & 24.2 & 24.9 & 32.7 & 38.5 \\
\midrule
From Scratch                   & \cmark  & 13.0 & 22.5 & 27.5 & 32.0 & 38.0 & 23.0 & 31.0 & 36.0 & 17.5 & 4.0  & 1.5  & 4.5  & 8.0  & 21.8 & 18.0 & 19.9 \\
Sparse Upcycling               & \xmark  & 33.5 & 56.0 & 60.5 & 65.0 & 48.0 & 30.5 & 39.5 & 52.5 & 26.5 & 26.5 & 9.0  & 15.0 & 21.5 & 23.5 & 31.5 & 35.9 \\
BTX                            & \xmark  & 36.0 & 64.5 & 66.0 & 69.0 & 50.5 & 33.0 & 39.5 & 56.0 & 29.0 & 31.5 & 11.0 & 17.5 & 24.5 & 25.0 & 33.0 & 39.1 \\
NVIDIA Upcycling               & \cmark  & 36.0 & 64.0 & 65.5 & 68.5 & 50.0 & 33.0 & 39.0 & 55.5 & 28.5 & 30.5 & 10.8 & 17.0 & 24.0 & 24.8 & 32.8 & 38.7 \\
Drop-Upcycling                 & \xmark  & 35.0 & 63.5 & 66.0 & 69.0 & 49.5 & 32.5 & 39.0 & 55.0 & 28.0 & 29.5 & 10.5 & 16.5 & 23.5 & 24.5 & 32.5 & 38.3 \\
Drop-Upcycling-f               & \cmark  & 25.5 & 28.5 & 33.0 & 36.0 & 43.5 & 24.5 & 38.0 & 39.5 & 17.0 & 6.0  & 2.5  & 6.0  & 9.5  & 21.5 & 19.0 & 23.3 \\
\rowcolor{gray!15}
\textbf{\divmoe{}}             & \cmark & \textbf{37.5} & \textbf{65.5} & \textbf{67.5} & \textbf{70.0} & \textbf{52.5} & \textbf{34.5} & \textbf{42.0} & \textbf{56.5} & \textbf{30.0} & \textbf{34.5} & \textbf{13.5} & \textbf{20.0} & \textbf{27.0} & \textbf{26.0} & \textbf{34.5} & \textbf{40.8} \\
\bottomrule
\end{tabular}
}
\end{table}

\section{Full Open-Source MoE Comparison (8 Benchmarks)}
\label{app:opensource_moe_full}

\Cref{tab:opensource_moe_full} reports the full 8-benchmark comparison underlying the abridged \Cref{tab:opensource_moe} in the main paper, additionally including ARC-Easy. The 8-benchmark average (\divmoe{} 64.5, Moonlight-MoE 64.5) is the figure cited in the abstract.

\begin{table}[h]
\caption{Comparison with open-source MoE models on the full 8-benchmark set. Same conventions as \Cref{tab:opensource_moe}.}
\label{tab:opensource_moe_full}
\centering
\small
\resizebox{\textwidth}{!}{
\begin{tabular}{lc|cccc|cccc|c}
\toprule
\textbf{Model} & \textbf{Size} & \textbf{ARC-c} & \textbf{ARC-e} & \textbf{MMLU} & \textbf{TriviaQA} & \textbf{HumanE.} & \textbf{MBPP} & \textbf{GSM8K} & \textbf{MATH} & \textbf{Avg.} \\
\midrule
OLMoE                                  & 1B/7B    & 49.2          & \underline{76.9} & 51.9          & 60.4 & \textbf{51.8} & 61.2          & 45.5          & 23.9          & 52.6 \\
OLMoE + T\"ulu-3 SFT                   & 1B/7B    & 52.4          & 77.2          & 55.8          & 63.5 & \underline{53.5} & 62.8          & 56.8          & 31.2          & 56.7 \\
DeepSeek-V2-Lite                       & 2.4B/16B & 52.1          & 70.3          & 58.3          & 65.1 & 29.9          & 43.2          & 41.1          & 17.1          & 47.1 \\
NVIDIA Up. + T\"ulu-3 SFT              & 2.4B/5.4B& 56.3          & 73.8          & 64.2          & 67.5 & 45.8          & 54.2          & 66.5          & 38.8          & 58.4 \\
Moonlight-MoE                          & 2.4B/16B & \textbf{65.5} & \textbf{79.3} & \underline{70.0} & \underline{66.3} & 48.1 & \textbf{63.8} & \textbf{77.4} & \underline{45.3} & \underline{64.5} \\
\rowcolor{gray!15}
\textbf{\divmoe{}}                     & 3.6B/12B & \underline{63.1} & 75.0 & \textbf{71.5} & \textbf{73.3} & \underline{51.4} & \underline{60.0} & \underline{74.4} & \textbf{47.5} & \textbf{64.5} \\
\bottomrule
\end{tabular}
}
\end{table}

\section{Attention Layer Initialization Ablation}
\label{app:attention_init}

We compare three attention initialization strategies for the non-expert components (\Cref{eq:nonexpert}). Mean initialization---averaging the base model and all specialists---is the default.

\begin{table}[h]
\caption{Attention initialization ablation on Qwen3-1.7B-Base after 500B Stage-2 CPT. Avg. is over the 9 general-knowledge benchmarks for direct comparability with prior reports.}
\label{tab:attention_init}
\centering
\small
\begin{tabular}{lccc|c}
\toprule
\textbf{Initialization} & \textbf{ARC-c} & \textbf{MMLU} & \textbf{HellaSwag} & \textbf{Avg. (9 bm.)} \\
\midrule
Mean of base + all specialists (default) & 45.0 & 64.5 & 63.5 & 58.3 \\
Base model only                          & 44.5 & 64.0 & 63.2 & 57.9 \\
Random specialist selection              & 43.5 & 63.0 & 62.5 & 57.0 \\
\bottomrule
\end{tabular}
\end{table}

\paragraph{Key takeaways.}
The default mean initialization is the best of the three, with the base-only variant trailing by 0.4 points. Random per-layer specialist selection performs worst ($-$1.3). The modest spread between mean and base-only confirms that the primary driver of \divmoe{}'s advantage is the FFN expert initialization, not the attention layer; mean initialization contributes a small but consistent boost.

\section{Learning Dynamics}
\label{app:learning_main}

\begin{figure}[h]
\centering
\includegraphics[width=0.95\linewidth]{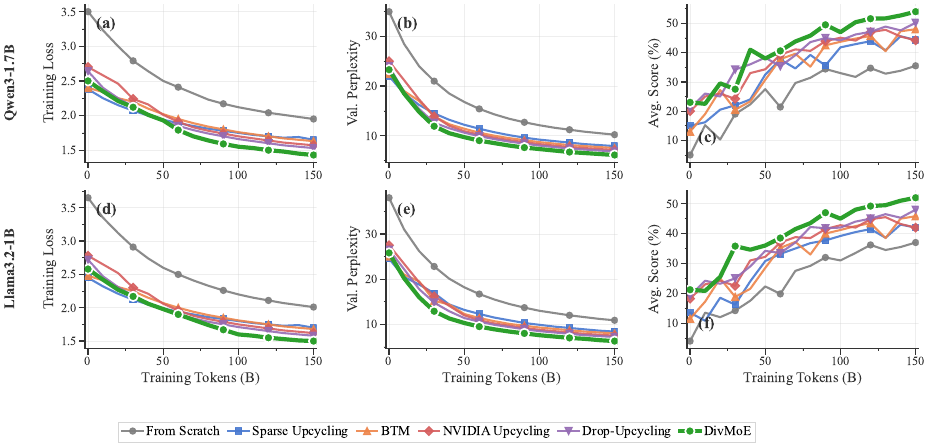}
\caption{Stage~2 learning dynamics. \divmoe{} achieves lower training loss and validation perplexity throughout training and converges to higher downstream performance, on both Qwen3-1.7B (top) and Llama3.2-1B (bottom).}
\label{fig:learning_curves}
\end{figure}

\Cref{fig:learning_curves} shows training loss, validation perplexity, and average downstream accuracy across Stage~2 CPT for Qwen3-1.7B-Base and Llama3.2-1B. \Cref{tab:convergence} extends the comparison to 300B and 500B tokens for the strongest baselines. Token-by-token tabulations are deferred to \Cref{app:learning_dynamics}.

\begin{table}[h]
\caption{Average accuracy (15 benchmarks) on Qwen3-1.7B-Base at extended checkpoints. \divmoe{}'s advantage is sustained: gaps narrow only mildly between 150B and 500B, indicating an architectural rather than initialization-only benefit.}
\label{tab:convergence}
\centering
\small
\begin{tabular}{l|ccc}
\toprule
\textbf{Method} & \textbf{@150B} & \textbf{@300B} & \textbf{@500B} \\
\midrule
NVIDIA Upcycling   & 47.5 & 49.5 & 51.1 \\
Drop-Upcycling     & 46.5 & 48.5 & 50.2 \\
\rowcolor{gray!15}
\divmoe{}          & \textbf{52.0} & \textbf{54.0} & \textbf{55.6} \\
\bottomrule
\end{tabular}
\end{table}

\paragraph{Sustained advantage.}
\divmoe{} leads at every checkpoint. The gap to NVIDIA Upcycling is 4.5 points at 150B and 4.5 at 500B; the absolute advantage is sustained throughout training, supporting the conclusion that the architectural benefits do not vanish with additional data.

\section{Extended Learning Dynamics}
\label{app:learning_dynamics}

Tables~\ref{tab:learning_qwen} and \ref{tab:learning_llama} report training loss, validation perplexity, and average downstream accuracy at 10B-token intervals during Stage~2 CPT. Best (lowest loss/PPL, highest accuracy) per checkpoint in \textbf{bold}.

\begin{table}[h]
\caption{Learning dynamics for Qwen3-1.7B-Base. Values reported at 10B-token intervals.}
\label{tab:learning_qwen}
\centering
\scriptsize
\setlength{\tabcolsep}{2.6pt}
\resizebox{\textwidth}{!}{
\begin{tabular}{l|cccccccccccccccc}
\toprule
\textbf{Tokens (B)} & 0 & 10 & 20 & 30 & 40 & 50 & 60 & 70 & 80 & 90 & 100 & 110 & 120 & 130 & 140 & 150 \\
\midrule
\multicolumn{17}{c}{\textit{Training Loss}} \\
From Scratch     & 3.50 & 3.23 & 3.00 & 2.79 & 2.64 & 2.50 & 2.41 & 2.32 & 2.23 & 2.17 & 2.12 & 2.08 & 2.04 & 2.01 & 1.98 & 1.95 \\
Sparse Upcycling & 2.38 & 2.25 & 2.15 & 2.07 & 2.00 & 1.94 & 1.89 & 1.85 & 1.81 & 1.78 & 1.75 & 1.72 & 1.70 & 1.68 & 1.69 & 1.65 \\
BTX              & 2.40 & 2.34 & 2.25 & 2.20 & 2.10 & 2.02 & 1.95 & 1.89 & 1.84 & 1.80 & 1.76 & 1.73 & 1.70 & 1.67 & 1.65 & 1.63 \\
NVIDIA Upcycling & 2.70 & 2.58 & 2.46 & 2.24 & 2.16 & 2.01 & 1.91 & 1.83 & 1.78 & 1.74 & 1.70 & 1.67 & 1.64 & 1.61 & 1.59 & 1.57 \\
Drop-Upcycling   & 2.64 & 2.40 & 2.25 & 2.13 & 2.02 & 1.92 & 1.85 & 1.79 & 1.74 & 1.70 & 1.66 & 1.63 & 1.60 & 1.57 & 1.55 & 1.53 \\
\textbf{\divmoe{}} & \textbf{2.50} & \textbf{2.35} & \textbf{2.21} & \textbf{2.12} & \textbf{2.01} & \textbf{1.93} & \textbf{1.79} & \textbf{1.70} & \textbf{1.64} & \textbf{1.59} & \textbf{1.55} & \textbf{1.53} & \textbf{1.50} & \textbf{1.48} & \textbf{1.45} & \textbf{1.43} \\
\midrule
\multicolumn{17}{c}{\textit{Validation Perplexity}} \\
From Scratch     & 35.0 & 28.5 & 24.0 & 21.0 & 18.5 & 16.8 & 15.4 & 14.3 & 13.4 & 12.7 & 12.1 & 11.6 & 11.2 & 10.8 & 10.5 & 10.2 \\
Sparse Upcycling & 22.0 & 18.5 & 16.2 & 14.5 & 13.2 & 12.2 & 11.4 & 10.7 & 10.1 & 9.6 & 9.2 & 8.9 & 8.6 & 8.3 & 8.1 & 7.9 \\
BTX              & 22.5 & 19.0 & 17.0 & 13.9 & 12.3 & 11.4 & 10.6 & 10.0 & 9.5 & 9.1 & 8.7 & 8.4 & 8.1 & 7.9 & 7.7 & 7.5 \\
NVIDIA Upcycling & 25.0 & 21.2 & 17.3 & 13.9 & 12.0 & 10.9 & 10.2 & 9.6 & 9.1 & 8.7 & 8.3 & 8.0 & 7.7 & 7.5 & 7.3 & 7.1 \\
Drop-Upcycling   & 24.0 & 18.8 & 15.4 & 12.6 & 11.5 & 10.6 & 9.9 & 9.3 & 8.8 & 8.4 & 8.0 & 7.7 & 7.4 & 7.2 & 7.0 & 6.8 \\
\textbf{\divmoe{}} & \textbf{23.3} & \textbf{18.5} & \textbf{14.8} & \textbf{11.9} & \textbf{10.5} & \textbf{9.7} & \textbf{9.0} & \textbf{8.5} & \textbf{8.0} & \textbf{7.6} & \textbf{7.3} & \textbf{7.0} & \textbf{6.7} & \textbf{6.5} & \textbf{6.3} & \textbf{6.1} \\
\midrule
\multicolumn{17}{c}{\textit{Average Downstream Accuracy (\%)}} \\
From Scratch     & 5.0 & 15.2 & 10.3 & 19.0 & 22.3 & 27.5 & 21.4 & 29.6 & 31.4 & 34.4 & 33.0 & 31.7 & 34.7 & 32.8 & 33.8 & 35.5 \\
Sparse Upcycling & 15.0 & 16.2 & 20.5 & 22.0 & 24.0 & 32.5 & 37.7 & 34.6 & 39.2 & 35.6 & 41.8 & 42.9 & 43.9 & 40.8 & 45.6 & 44.3 \\
BTX              & 13.0 & 19.0 & 27.2 & 20.5 & 23.3 & 30.7 & 37.8 & 39.6 & 35.2 & 42.5 & 43.7 & 44.8 & 45.7 & 40.5 & 47.3 & 48.0 \\
NVIDIA Upcycling & 20.0 & 25.0 & 26.0 & 24.2 & 33.0 & 34.3 & 39.3 & 41.1 & 40.6 & 43.9 & 45.1 & 44.1 & 47.0 & 47.8 & 45.5 & 44.2 \\
Drop-Upcycling   & 21.0 & 26.0 & 25.0 & 34.2 & 36.0 & 38.3 & 35.3 & 39.1 & 43.6 & 45.0 & 44.2 & 46.2 & 47.1 & 48.9 & 47.6 & 50.2 \\
\textbf{\divmoe{}} & \textbf{23.0} & 22.5 & \textbf{29.5} & 27.5 & \textbf{41.0} & 38.0 & \textbf{40.6} & \textbf{43.8} & \textbf{45.8} & \textbf{49.5} & \textbf{47.0} & \textbf{50.4} & \textbf{51.6} & \textbf{51.7} & \textbf{52.7} & \textbf{54.0} \\
\bottomrule
\end{tabular}
}
\end{table}

\begin{table}[h]
\caption{Learning dynamics for Llama3.2-1B. Values reported at 10B-token intervals.}
\label{tab:learning_llama}
\centering
\scriptsize
\setlength{\tabcolsep}{2.6pt}
\resizebox{\textwidth}{!}{
\begin{tabular}{l|cccccccccccccccc}
\toprule
\textbf{Tokens (B)} & 0 & 10 & 20 & 30 & 40 & 50 & 60 & 70 & 80 & 90 & 100 & 110 & 120 & 130 & 140 & 150 \\
\midrule
\multicolumn{17}{c}{\textit{Training Loss}} \\
From Scratch     & 3.65 & 3.35 & 3.12 & 2.91 & 2.74 & 2.60 & 2.50 & 2.41 & 2.33 & 2.26 & 2.20 & 2.15 & 2.11 & 2.07 & 2.04 & 2.01 \\
Sparse Upcycling & 2.45 & 2.32 & 2.21 & 2.12 & 2.05 & 1.99 & 1.94 & 1.90 & 1.86 & 1.83 & 1.80 & 1.77 & 1.75 & 1.73 & 1.74 & 1.70 \\
BTX              & 2.48 & 2.41 & 2.31 & 2.25 & 2.15 & 2.07 & 2.00 & 1.94 & 1.89 & 1.85 & 1.81 & 1.78 & 1.75 & 1.72 & 1.70 & 1.68 \\
NVIDIA Upcycling & 2.78 & 2.65 & 2.52 & 2.30 & 2.21 & 2.06 & 1.96 & 1.88 & 1.83 & 1.79 & 1.75 & 1.72 & 1.69 & 1.66 & 1.64 & 1.62 \\
Drop-Upcycling   & 2.72 & 2.47 & 2.31 & 2.18 & 2.07 & 1.97 & 1.90 & 1.84 & 1.79 & 1.75 & 1.71 & 1.68 & 1.65 & 1.62 & 1.60 & 1.58 \\
\textbf{\divmoe{}} & \textbf{2.58} & \textbf{2.42} & \textbf{2.27} & \textbf{2.17} & \textbf{2.06} & \textbf{1.98} & \textbf{1.90} & \textbf{1.82} & \textbf{1.74} & \textbf{1.67} & \textbf{1.60} & \textbf{1.58} & \textbf{1.55} & \textbf{1.53} & \textbf{1.51} & \textbf{1.50} \\
\midrule
\multicolumn{17}{c}{\textit{Validation Perplexity}} \\
From Scratch     & 38.0 & 31.0 & 26.2 & 22.8 & 20.1 & 18.2 & 16.7 & 15.5 & 14.5 & 13.7 & 13.0 & 12.5 & 12.0 & 11.6 & 11.2 & 10.9 \\
Sparse Upcycling & 24.5 & 20.5 & 18.8 & 16.9 & 14.4 & 13.2 & 12.3 & 11.5 & 10.9 & 10.3 & 9.9 & 9.5 & 9.2 & 8.9 & 8.6 & 8.4 \\
BTX              & 25.0 & 21.0 & 19.5 & 15.2 & 14.4 & 12.3 & 11.4 & 10.7 & 10.1 & 9.7 & 9.3 & 9.0 & 8.7 & 8.4 & 8.2 & 8.0 \\
NVIDIA Upcycling & 27.5 & 23.4 & 19.3 & 16.2 & 14.1 & 12.0 & 10.9 & 10.2 & 9.7 & 9.2 & 8.8 & 8.5 & 8.2 & 7.9 & 7.7 & 7.5 \\
Drop-Upcycling   & 26.5 & 22.2 & 17.8 & 14.8 & 12.9 & 11.4 & 10.6 & 9.9 & 9.4 & 8.9 & 8.5 & 8.2 & 7.9 & 7.6 & 7.4 & 7.2 \\
\textbf{\divmoe{}} & \textbf{25.8} & \textbf{20.2} & \textbf{16.0} & \textbf{12.9} & \textbf{11.3} & \textbf{10.3} & \textbf{9.5} & \textbf{8.9} & \textbf{8.4} & \textbf{8.0} & \textbf{7.6} & \textbf{7.3} & \textbf{7.0} & \textbf{6.8} & \textbf{6.5} & \textbf{6.3} \\
\bottomrule
\end{tabular}
}
\end{table}

\section{Granularity Ablation Details}
\label{app:granularity}

\Cref{tab:granularity_app} gives the full numerical breakdown of \Cref{fig:granularity}, on Qwen3-1.7B-Base.

\begin{table}[h]
\caption{Effect of expert granularity on training dynamics. $m=k=1$ corresponds to coarse-grained MoE (4 experts), $m=k=2$ is the default \divmoe{} (8 experts), and $m=k=4$ represents finer-grained MoE (16 experts).}
\label{tab:granularity_app}
\centering
\scriptsize
\setlength{\tabcolsep}{2.6pt}
\resizebox{\textwidth}{!}{
\begin{tabular}{l|cccccccccccccccc}
\toprule
\textbf{Tokens (B)} & 0 & 10 & 20 & 30 & 40 & 50 & 60 & 70 & 80 & 90 & 100 & 110 & 120 & 130 & 140 & 150 \\
\midrule
\multicolumn{17}{c}{\textit{Training Loss}} \\
$m{=}k{=}1$ & 2.80 & 2.52 & 2.35 & 2.22 & 2.12 & 2.04 & 1.97 & 1.91 & 1.86 & 1.82 & 1.78 & 1.75 & 1.72 & 1.69 & 1.67 & 1.65 \\
$m{=}k{=}4$ & 2.97 & 2.63 & 2.48 & 2.21 & 2.07 & 1.97 & 1.87 & 1.81 & 1.76 & 1.71 & 1.67 & 1.64 & 1.61 & 1.58 & 1.56 & 1.54 \\
$\bm{m{=}k{=}2}$ & 2.90 & 2.58 & 2.36 & 2.15 & 2.01 & 1.94 & 1.85 & 1.76 & 1.71 & 1.66 & 1.62 & 1.59 & 1.56 & 1.53 & 1.51 & \textbf{1.49} \\
\midrule
\multicolumn{17}{c}{\textit{Validation Perplexity}} \\
$m{=}k{=}1$ & 16.5 & 12.8 & 10.9 & 9.6 & 8.7 & 8.0 & 7.5 & 7.1 & 6.7 & 6.4 & 6.2 & 6.0 & 5.8 & 5.6 & 5.5 & 5.4 \\
$m{=}k{=}4$ & 18.9 & 15.2 & 12.1 & 9.8 & 8.2 & 7.2 & 6.7 & 6.3 & 5.9 & 5.6 & 5.4 & 5.2 & 5.0 & 4.8 & 4.7 & 4.6 \\
$\bm{m{=}k{=}2}$ & 17.8 & 14.8 & 11.6 & 9.3 & 8.1 & 6.7 & 6.2 & 5.8 & 5.4 & 5.1 & 4.9 & 4.7 & 4.5 & 4.4 & 4.3 & \textbf{4.2} \\
\midrule
\multicolumn{17}{c}{\textit{MMLU Accuracy (\%)}} \\
$m{=}k{=}1$ & 20.0 & 24.2 & 25.5 & 29.0 & 30.5 & 32.0 & 31.5 & 35.0 & 36.3 & 35.5 & 38.5 & 39.5 & 40.3 & 41.0 & 41.6 & 42.1 \\
$m{=}k{=}4$ & 14.3 & 17.0 & 23.2 & 29.5 & 32.5 & 36.0 & 34.0 & 39.8 & 41.3 & 40.7 & 42.9 & 44.9 & 45.8 & 45.5 & 47.1 & 47.6 \\
$\bm{m{=}k{=}2}$ & 16.2 & 17.5 & 25.2 & 34.0 & 33.8 & 38.5 & 37.0 & 42.8 & 44.5 & 43.0 & 45.3 & 48.4 & 49.3 & 49.1 & 50.7 & \textbf{51.2} \\
\midrule
\multicolumn{17}{c}{\textit{HellaSwag Accuracy (\%)}} \\
$m{=}k{=}1$ & 20.0 & 24.5 & 22.5 & 27.0 & 30.0 & 33.5 & 31.8 & 36.8 & 36.5 & 40.0 & 42.3 & 42.5 & 44.5 & 43.4 & 45.2 & 45.9 \\
$m{=}k{=}4$ & 16.4 & 22.0 & 20.5 & 26.0 & 34.0 & 37.2 & 36.0 & 38.3 & 40.3 & 42.0 & 43.5 & 43.8 & 45.9 & 45.8 & 47.6 & 48.3 \\
$\bm{m{=}k{=}2}$ & 18.5 & 23.2 & 22.5 & 30.8 & 35.2 & 35.8 & 41.8 & 42.3 & 42.5 & 44.3 & 45.9 & 46.3 & 47.5 & 47.5 & 48.3 & \textbf{49.0} \\
\bottomrule
\end{tabular}
}
\end{table}

\paragraph{Key observations.}
(1)~$m{=}k{=}2$ achieves the best trade-off between training stability (lower initial loss than $m{=}k{=}4$) and final accuracy.
(2)~$m{=}k{=}1$ converges fastest initially but plateaus at lower accuracy due to limited expert capacity.
(3)~The granularity gap widens on harder benchmarks (MMLU 9.1\% vs.\ HellaSwag 3.1\% at 150B), suggesting fine-grained experts benefit knowledge-intensive tasks the most.

\end{document}